\documentclass{article}

\usepackage{arxiv}

\usepackage[utf8]{inputenc} %
\usepackage[T1]{fontenc}    %
\usepackage{hyperref}       %
\usepackage{url}            %
\usepackage{booktabs}       %
\usepackage{amsfonts}       %
\usepackage{nicefrac}       %
\usepackage{microtype}      %
\usepackage{cleveref}       %
\usepackage{lipsum}         %
\usepackage{graphicx}
\usepackage{natbib}
\usepackage{doi}
\usepackage{wrapfig}

\title{LPML: LLM-Prompting Markup Language for Mathematical Reasoning}

\author{Ryutaro Yamauchi\thanks{This work was done when the author was at Albert Inc.} \\ The University of Tokyo \\ ryutaro\_yamauchi@weblab.t.u-tokyo.ac.jp
        \And
        Sho Sonoda \\ Center for Advanced Intelligence Project, RIKEN \\ sho.sonoda@riken.jp
        \AND
        Akiyoshi Sannai \\ Kyoto University \\ sannai.akiyoshi.7z@kyoto-u.ac.jp
        \And
        Wataru Kumagai \\ The University of Tokyo \\ kumagai@weblab.t.u-tokyo.ac.jp
        }

\begin{document}
\maketitle
\begin{abstract}

In utilizing large language models (LLMs) for mathematical reasoning, addressing the errors in the reasoning and calculation present in the generated text by LLMs is a crucial challenge. In this paper, we propose a novel framework that integrates the Chain-of-Thought (CoT) method with an external tool (Python REPL). We discovered that by prompting LLMs to generate structured text in XML-like markup language, we could seamlessly integrate CoT and the external tool and control the undesired behaviors of LLMs. With our approach, LLMs can utilize Python computation to rectify errors within CoT. We applied our method to ChatGPT (GPT-3.5) to solve challenging mathematical problems and demonstrated that combining CoT and Python REPL through the markup language enhances the reasoning capability of LLMs. Our approach enables LLMs to write the markup language and perform advanced mathematical reasoning using only zero-shot prompting.
\end{abstract}

\section{Introduction}

Large language models (LLMs) have demonstrated remarkable success in a wide range of natural language processing (NLP) tasks, despite being trained on the simplistic task of predicting subsequent words in sentences. LLMs possess the capability to extract rules and principles from the provided text (called In-Context Learning). This enables LLMs to solve even novel tasks by being \textit{prompted} with a few-shot example~\citep{NEURIPS2020_1457c0d6}. Furthermore, it has been reported that effective prompts can improve the performance of LLMs. For instance, we can improve the ability of LLMs to perform arithmetic reasoning by prompting them to output not only the final answer but also intermediate steps involved in problem-solving. This technique, reffered to as Chain-of-Thought (CoT), is widely used ~\citep{NEURIPS2022_9d560961}.

One of the most essential challenges in employing LLMs for mathematical reasoning is the management of calculation and reasoning errors in LLMs' outputs~\citep{hendrycksmath2021}.
One approach is to improve the performance of LLMs by increasing the model size~\citep{openai2023gpt4} or by fine-tuning with mathematical and technical texts~\citep{NEURIPS2022_18abbeef, chung2022scaling}, but training LLMs are often expensive.
As a result, alternative strategies that improve performance without modifying the LLMs themselves have been sought.
Several studies have LLMs generate explicit reasoning step~\citep{NEURIPS2022_9d560961, zhou2023leasttomost, nye2021work}.
\citep{wang2023selfconsistency} proposes a \textit{self-consistency} strategy to have LLMs generate multiple reasoning paths and merge them.
Furthermore, methods for integrating LLMs with external tools have been proposed~\citep{gao2022pal, yao2022react, schick2023toolformer, chen2022program, mishra-etal-2022-lila}. Given LLMs' limitations with large-number calculations and complex reasoning, outsourcing these tasks to external tools can potentially reduce reasoning errors. For instance, PAL~\citep{gao2022pal} outsources calculations that would be difficult for LLMs alone to a computer by few-shot prompting with pairs of a problem and a Python code to solve it. However, these methods treat the use of external tools as a module independent of other reasoning processes, so the accuracy limitations of each module constrain overall performance. To fundamentally improve the accuracy of reasoning, the consistency of the external tool's behavior should be monitored by LLMs themselves.

\begin{wrapfigure}{r}[0pt]{0.5\textwidth}
\centering
\vspace{-4mm}
\includegraphics[width=1.0\linewidth]{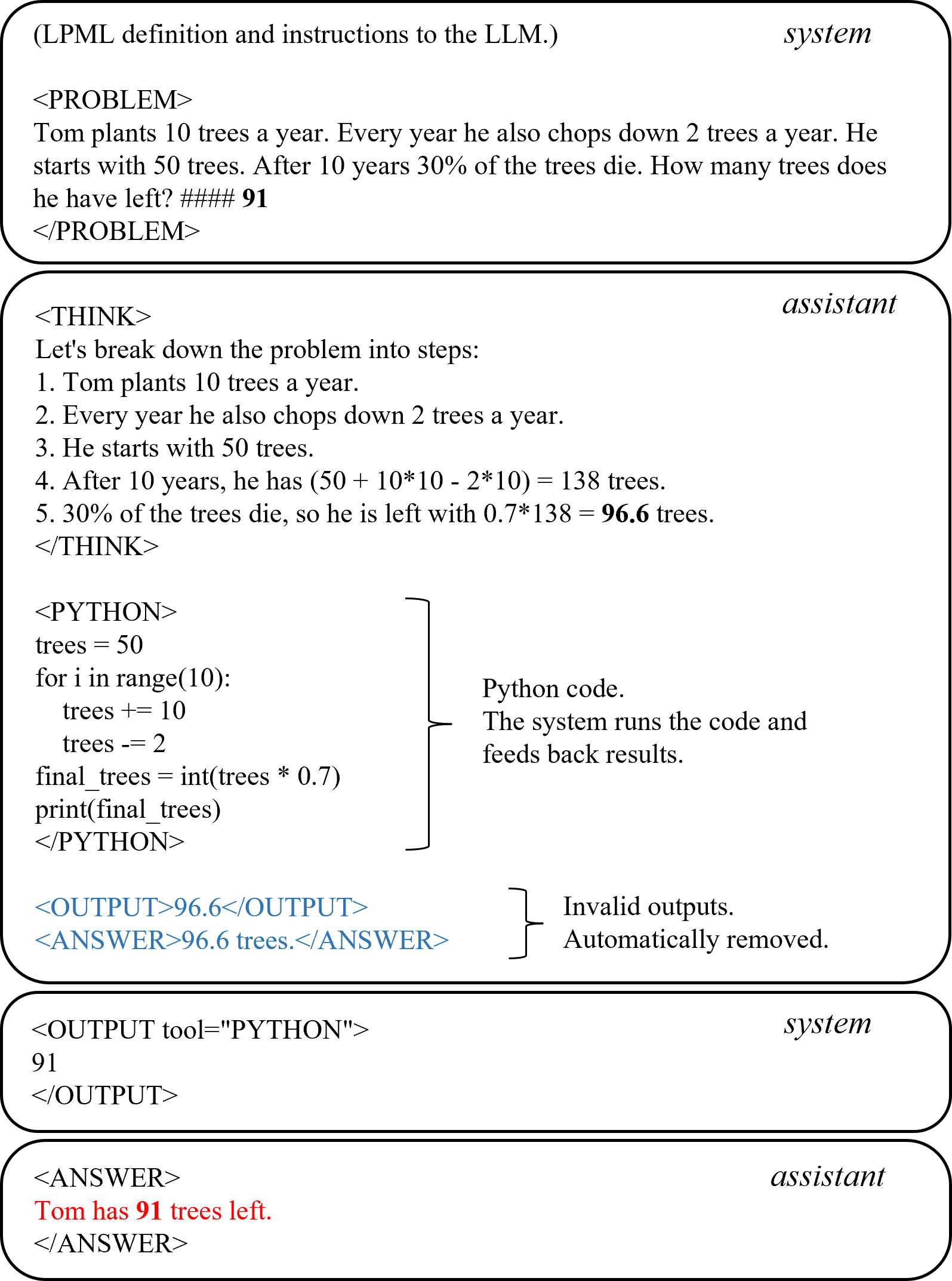}
\caption{An example of the reasoning process: \textit{assistant} (LLM) outputs CoT and Python code, while the \textit{system} (computer) provides feedback on the code's execution results. Since \textit{assistant} may output invalid elements, \textit{system} removes them.}
\label{fig:chat}
\vspace{-4.5mm}
\end{wrapfigure}

\begin{wrapfigure}{r}[0pt]{0.59\textwidth}
\centering
\vspace{-4mm}
\includegraphics[width=1.0\linewidth]{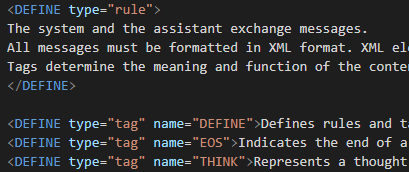}
\caption{The prompt: the syntax definition of LPML and instructions on how to solve the problem written in LPML.}
\label{fig:xml}
\vspace{-4.5mm}
\end{wrapfigure}

In this paper, we propose a novel framework for integrating the Python REPL (read–eval–print loop) with LLMs.
Our approach differs from previous works in that we use Python code (and its execution results) as part of the CoT rather than having LLMs directly generate the Python code to solve the problem.
Specifically, we construct an interaction loop between LLMs and the Python REPL by having LLMs generate both the CoT and the Python code and then feeding the results of the code execution back to LLMs to induce further CoTs (Fig.~\ref{fig:overview}). This allows LLMs to have two reasoning paths, CoT and Python code execution and is expected to improve the reasoning performance by self-consistency.
However, we found that simply feeding back Python code execution results does not lead LLMs to behave as we expected: when the code is written in the CoT, LLMs do not wait for the execution results but output dummy results as well. Also, when LLMs make mistakes in CoT reasoning, they tend to think that the code is incorrect, even if they write correct Python code and get the result of that execution.
To avoid these problems, we define an XML-like markup language (LLM-prompting markup language, LPML) and feed its syntax to LLMs to prompt them to output text structured in the markup language (Fig.~\ref{fig:xml}). LPML includes \verb|THINK| tags for CoT, \verb|PYTHON| tags for Python code, and \verb|OUTPUT| tags for the results of Python code execution, etc. By using these tags, the LLMs' output text is structured.
The advantages of this approach are as follows: 1. we can easily parse the text generated by LLMs and segment it by function and meaning; 2. we can mechanically remove invalid text generated by LLMs, such as dummy execution results because the tags correspond to their contents; and 3. we can define relationships between tags, which enables us to make LLMs trust the results of Python calculations over those in the CoT, which are prone to errors. As a result, we achieve the \textit{fall-back in reasoning} that LLMs often have difficulty with.
We applied our method to ChatGPT~\citep{ChatGPT} to solve challenging mathematical problems and demonstrated that the combination of CoT and Python computation improves LLMs' reasoning ability.

\section{Methods}

\subsection{Interaction loop between ChatGPT and Python REPL}

In this study, we used OpenAI ChatGPT (GPT-3.5-Turbo)~\citep{ChatGPT} via API, which is an LLM that takes a sequence of messages $[m_0, \cdots, m_k]$ and predicts the next message $m_{k+1}$. The API specification allows us to set three roles as the speaker of a message: \textit{assistant}, \textit{user}, and \textit{system}, where \textit{assistant} is an AI, \textit{user} is a human, and \textit{system} is a conversation context manager.

\begin{figure*}[t]
\begin{center}
   \includegraphics[width=1.0\linewidth]{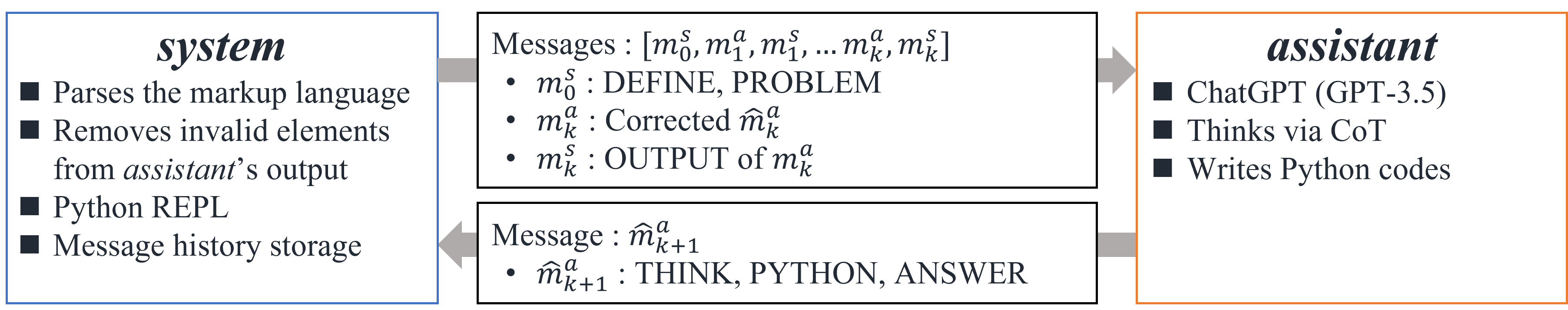}
\end{center}
   \caption{An overview of the mathematical reasoning process. \textit{system} and \textit{assistant} continue to interact until an answer is obtained. All messages are structured in the markup language.}
\label{fig:overview}
\end{figure*}

As shown below, we designed the mathematical reasoning process as an interaction between \textit{system} and \textit{assistant}. First, as \textit{system}, we present the problem to be solved and define the syntax of the markup language and rules of reasoning ($m^s_0$). Then, when \textit{assistant} returns an output ($\hat{m}^a_k$), \textit{system} analyzes it, and if it contains Python code, \textit{system} returns the execution result ($m^s_k$). If \textit{assistant} outputs any undesired elements, such as fake Python execution results, \textit{system} removes them from the \textit{assistant}'s output ($m^a_k$). We also remove the answer if \textit{assistant} outputs both the Python code and the answer in a single message, and then feed back the Python code execution results. We made \textit{system} and \textit{assistant} repeat this process until \textit{assistant} outputs the answer (Fig.~\ref{fig:overview}).

\subsection{LLM-prompting markup language}

The LPML, the markup language we have defined, is based on XML syntax and is a set of elements consisting of content enclosed in a start tag \verb|<TAG>| and end tag \verb|</TAG>|. 
Since the dataset used to train LLMs is assumed to include text written in XML and HTML, LLMs can write LPML that is similar to them. 
The LPML includes the tags \verb|DEFINE|, \verb|PROBLEM|, \verb|ANSWER|, \verb|THINK|, \verb|PYTHON|, and \verb|OUTPUT|. \verb|DEFINE| tag is for defining tags and rules. The syntax of the language and the tags are defined by \verb|DEFINE| tag. \verb|PROBLEM| and \verb|ANSWER| tags are for describing the problem and answer, and the solving process ends when \textit{assistant} outputs \verb|ANSWER| tag. \verb|THINK| tag is for describing thoughts. By defining \verb|THINK| tag as ``a tag for describing thinking step-by-step'', we induce Zero-shot-CoT~\citep{NEURIPS2022_8bb0d291}. \verb|PYTHON| tag is for describing Python code. When \textit{assistant} uses \verb|PYTHON| tag, \textit{system} returns the execution result using \verb|OUTPUT| tag. By using these tags, all messages, including the initial system prompt, are written in the LPML, thus strongly conditioning \textit{assistant} to output in the LPML.

\subsection{Instructions to ChatGPT}
In the initial prompt ($m_0^s$), we instruct ChatGPT to solve the math problem in the PROBLEM tag using the two tags \verb|PYTHON| and \verb|THINK| and to write the answer using the \verb|ANSWER| tag. We do not give a few-shot example of how to solve the problem but just give the LPML definition, instructions on which tags to use, and some notes.

We also instructed ChatGPT to trust the contents of \verb|OUTPUT| tags rather than the contents of \verb|THINK| tags. This enables LLMs to ignore errors in CoT by referring to the results of Python code execution. Without this instruction, ChatGPT would either assume that the Python code was incorrect and fall into a loop of repeated debugging or ignore the contents of the \verb|OUTPUT| tag.

\section{Experiments ans Results}

\subsection{Datasets and Evaluation}
We evaluated our method on the GSM8K dataset~\citep{cobbe2021training} and the MATH dataset~\citep{hendrycksmath2021}. Both datasets are available under MIT License.
The GSM8K is a dataset of 8.5K grade-school level math word problems, and we used the 1319 test problems in it for our evaluation.
The MATH dataset consists of 12,500 challenging competitive math problems, and the dataset contains seven subjects: Prealgebra, Algebra, Number Theory, Counting and Probability, Geometry, Intermediate Algebra, and Precalculus. We sampled 20 problems from each of the six subjects except Geometry and solved these 120 problems using our method.

We scored only the content of the \verb|ANSWER| tags in evaluating our method. And we manually scored the MATH dataset because of the arbitrariness of how to write the answers.

\subsection{Baselines}
We used Chain-of-Thought (CoT)~\citep{NEURIPS2022_9d560961} and Program-aided Language models (PAL)~\citep{gao2022pal} for comparison.
PAL is a method of solving mathematical problems by having LLMs write Python code. Since PAL uses Python code and does not use CoT, we can evaluate the effectiveness of our method using both Python and CoT by comparing our method with PAL.
We used ChatGPT (GPT-3.5-Turbo) as the backbone LLM for both methods.
Since both the GSM8K and MATH datasets contain manually written reasoning paths, we randomly selected five problems from each dataset and used them as few-shot examples for CoT.
For PAL, we used the author's implementation. The author's implementation contains Python code examples for few-shot prompting. Since this example is not tuned for the MATH dataset, we recreated the prompt using problems from the MATH dataset, but since there was no significant difference in performance, we present results from the author's prompt in subsequent reports.

\subsection{Implementation details}
Our implementation of \textit{system} behaves as follows. First, \textit{system} parses the output of \textit{assistant} $\hat{m}_k^a$ and creates a syntax tree. Next, it removes invalid elements from the syntax tree, specifically tags that must not be used by \textit{assistant} and text that is not enclosed in tags. Also, if the syntax tree contains both \verb|PYTHON| and \verb|ANSWER| elements, the \verb|ANSWER| element is removed. As a result, a normalized \textit{assistant} output $m_k^a$ is generated. Then, \textit{system} runs all \verb|PYTHON| elements in the syntax tree using the Python REPL and generates the system message $m_k^s$, which is a collection of standard/error outputs of the codes. Finally, system gives the message history $[m_0^s, m_1^a, m_1^s, \cdots, m_k^a, m_k^s]$ to \textit{assisntant}.

We set the number of \textit{system} and \textit{assistant} interactions to 8 turns, excluding the initial prompt. If the normalized \textit{assistant} output $m_k^a$ contains an \verb|ANSWER| tag, the interaction is terminated at that point. If no \verb|ANSWER| tag was output once throughout the interaction, the interaction is redone from the beginning. We set the ChatGPT temperature to 0. However, if the ANSWER tag was not output after 8 interactions, we increased the temperature in 0.1 increments for up to 5 attempts.

\subsection{Results}

Table~\ref{table:result} shows the comparison results between the proposed method and the previous method on the GSM8K and MATH datasets. And Figure~\ref{fig:math_result} shows the success rate of each method in each subject of the MATH data set. In the GSM8K dataset, PAL showed a higher accuracy rate than our method. However, our method showed a higher accuracy rate in the MATH dataset. The problems included in the MATH dataset are relatively difficult, and it is not always possible to implement Python code to solve them directly. Our method can combine the strengths of Python calculations and CoT reasonings, enabling LLMs to use CoT to solve reasoning that cannot be achieved with Python code while reducing CoT reasoning errors with Python calculations. On the other hand, the problems in the GSM8K dataset can be easily implemented with Python code, which may have resulted in decreased performance due to CoT-derived reasoning errors.

\makeatletter
\newcommand{\figcaption}[1]{\def\@captype{figure}\caption{#1}}
\newcommand{\tblcaption}[1]{\def\@captype{table}\caption{#1}}
\makeatother

\begin{figure}[h]
{\renewcommand{\baselinestretch}{0.95}
\def\@captype{table}
\begin{minipage}[t]{.48\textwidth}
\begin{center}
\scalebox{1}{\tabcolsep = 15pt
\begin{tabular}{lcc}

\hline
\hline
     & GSM8K & MATH   \\ \hline
CoT  & 57.1  & 31.7 \\
PAL  & \textbf{79.8}  & 47.5 \\
Ours & 76.6  & \textbf{60.0} \\ \hline
\hline

\end{tabular}
}
\end{center}
\tblcaption{Comparison of the success rate of each method on the GSM8K and MATH.}
\label{table:result}
\end{minipage}
}
\hfill
\begin{minipage}[c]{.48\textwidth}

\includegraphics[width=1.0\linewidth]{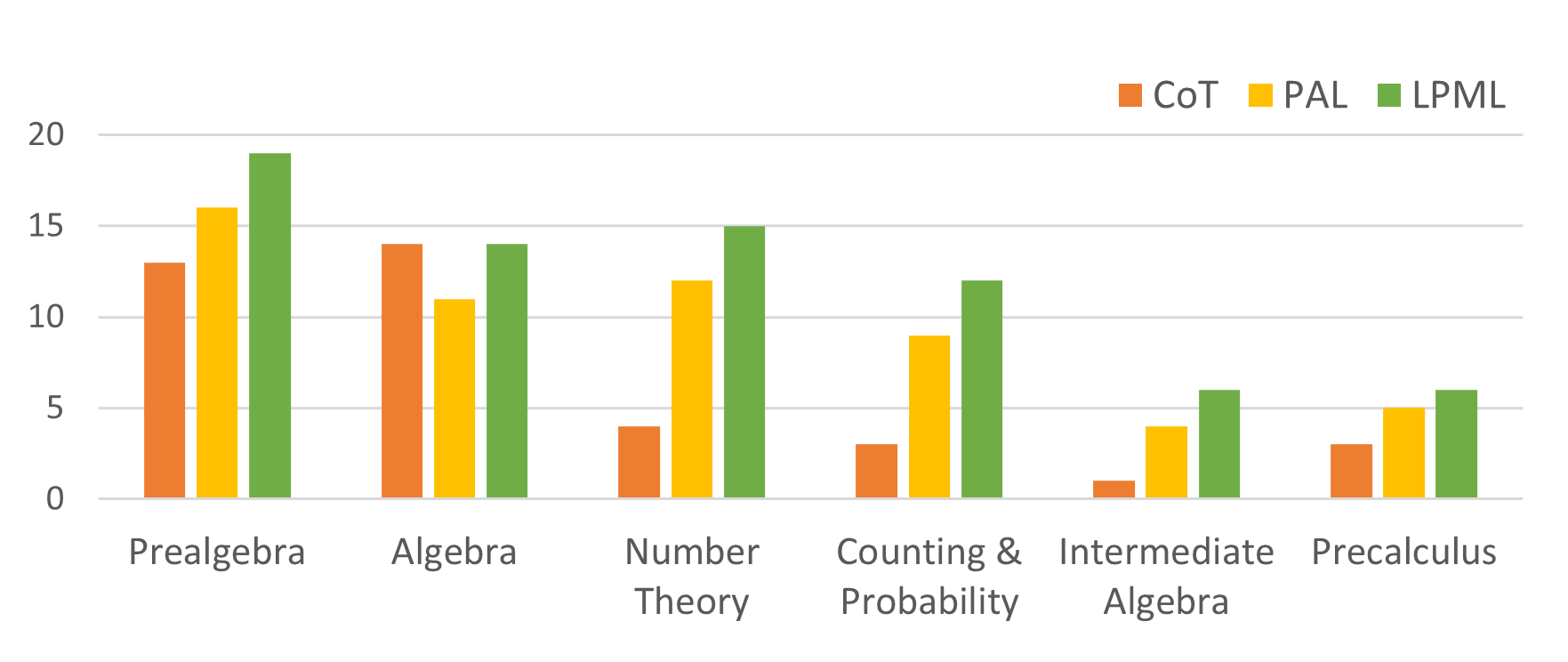}

\caption{The number of correct answers per subject in the MATH dataset.}
\label{fig:math_result}
\end{minipage}
\end{figure}

\section{Discussion and Conclusion}
We demonstrate that the reasoning ability of LLMs can be enhanced by having LLMs generate structured text in a markup language.  
Structuring the text generated by LLMs allows us to easily parse it, allowing us to control the behavior of LLMs and integrate LLMs with the Python REPL flexibly.  
Also, the tags qualify and divide texts into sections, allowing us to define relationships between texts and inducing LLMs behavior that prioritizes certain sentences over others. 

In addition to speech, humans have a variety of language output ports, including internal speech and writing. The use of markup languages to qualify and divide generated text into semantic and functional units is the equivalent to LLMs having such a variety of language output ports. In this study, two tags, \verb|PYTHON| and \verb|THINK|, were given to LLMs, but theoretically, any variety of tags can be defined, such as tags that output queries for web search or database operations or tags that output internal speech. However, the use of such a markup language is not a natural behavior for existing LLMs, and it is unclear whether LLMs can handle more complex syntax. Therefore, one intriguing future research direction is the fine-tuning of LLMs based on the use of markup languages.

\section{Limitations}
While we have successfully corrected some CoT errors with Python execution result feedback, this approach does not guarantee consistent success. Consequently, the number of errors caused by the limitations of LLMs will be higher compared to scenarios where LLMs generate Python code without CoT. As a result, in situations where there is no synergy between CoT and external tools, the performance will be reduced. Additionally, while it is possible to make LLMs acknowledge their errors, understanding the specific details of these errors often remains challenging. We are currently exploring potential solutions to address these limitations in principle.

Because LLMs are trained on a large corpus of text collected from the web, they have the potential to amplify biases in the training data. Thus, our method, which utilizes the statistical patterns that LLMs acquire via prompting, has the same problem. However, we believe that our method enables the seamless integration of LLMs with external resources and has the potential to correct the biases of LLMs.

\bibliography{references}
\bibliographystyle{unsrtnat}

\appendix

\section{Example Prompt}
\label{sec:appendix1}
Figure~\ref{fig:prompt} shows an example of the prompts we used. By writing the prompt itself in LPML format, we can strongly prompt LLMs to generate text in LPML format.

\section{Examples of Reasoning Processes}
\label{sec:appendix2}
Figures~\ref{fig:chat1}, \ref{fig:chat2}, and \ref{fig:chat3} show examples of the inference process.
LPML definitions and instructions to LLMs are omitted. In the example in Figure~\ref{fig:chat1}, the LLM makes a computational error within the CoT, but corrects the error using the Python results and reaches the correct answer. On the other hand, in the example in Figure~\ref{fig:chat2}, the LLM writes the correct Python code but ignores the results and outputs a different answer. In the example in Figure~\ref{fig:chat3}, there is a problem with the Python code, but the CoT has arrived at the correct answer. As these examples show, LLM's way to use external tools is inconsistent, and the success or failure of the reasoning still depends on chance.

\begin{figure*}[h]
\begin{center}
{\scriptsize
\rule{\textwidth}{1.2pt}
\begin{verbatim}
<DEFINE type="rule">
The system and the assistant exchange messages.
All messages must be formatted in XML format. XML element ::= <tag>content</tag> | <tag>
Tags determine the meaning and function of the content. The content must not contradict the definition of the tag.
</DEFINE>

<DEFINE type="tag" name="DEFINE">Defines rules and tags. The defined content is absolute.</DEFINE>
<DEFINE type="tag" name="EOS">Indicates the end of a message.</DEFINE>
<DEFINE type="tag" name="THINK">Represents a thought process. The thought process must be described step by step.</DEFINE>
<DEFINE type="tag" name="PYTHON">Represents an executable Python code.</DEFINE>
<DEFINE type="tag" name="OUTPUT">Represents a messages from the system to the assistant.</DEFINE>
<DEFINE type="tag" name="PROBLEM">Represents a problem to be solved.</DEFINE>
<DEFINE type="tag" name="ANSWER">Represents an answer to the problem.</DEFINE>

<DEFINE type="rule" role="system">
The system is a computer that supports the assistant.
When the system finds a PYTHON element in the assistant's message, the system executes it and returns the result to the 
assistant using the OUTPUT tag.
</DEFINE>

<DEFINE type="rule" role="assistant">
The assistant solves mathematical problems using the PYTHON tag and writes the answer using the ANSWER tag. The assistant is 
only allowed to use the PYTHON, THINK, and ANSWER tags.
The assistant translates the given problem into Python code to solve it and resolves the problem by executing it through the
system.
When performing calculations, algebraic manipulations, or reasoning, the assistant must always use the PYTHON tag. In other
words, calculations should not be performed within the THINK tag. If calculations are unavoidably performed within the THINK 
tag, the assistant must restart all those calculations from the beginning using the PYTHON tag. The assistant should never 
write the answer within the THINK tag.
The assistant must trust calculations performed using the PYTHON tag more than those performed within the THINK tag. 
Therefore, if the content of the OUTPUT element differs from the content of the THINK element, the assistant must assume there 
is an error in the THINK element and correct it. After correcting the content of the THINK element, the assistant must 
validate it again using the PYTHON tag. If 
no matter how many corrections are made to the content of the THINK element, it still contradicts the content of the OUTPUT 
element, the assistant must consider the OUTPUT element to be correct.
If the assistant uses the THINK element, it can use the ANSWER tag only when the content of the THINK element and the content 
of the OUTPUT element are not contradictory. However, if modifying the content of the THINK element twice still results in 
conflicting content between the two, the assistant must consider the content of the OUTPUT element to be correct.
When using the PYTHON tag, the assistant must use the `print()` function to output values. Additionally, after using the 
PYTHON tag, the 
assistant must immediately use the EOS tag to interrupt the message and wait for the system to return the execution result. In 
other words, it should be done as follows: <PYTHON>code</PYTHON><EOS>
The assistant can use the following Python libraries: sympy, math, numpy.
It is not necessary to solve problems analytically, and the use of brute force or numerical methods with the PYTHON tag is 
also allowed.
</DEFINE>

<PROBLEM>
An integer is randomly selected from the integers 0 through 99, inclusive. What is the probability that the integer is in the
range 10 through 20, inclusive? Express your answer as a common fraction.
</PROBLEM>

<OUTPUT>
This is the system. Waiting for a message from the assistant.
</OUTPUT>

<EOS>
\end{verbatim}
\vspace{2mm}
\rule{\textwidth}{1.2pt}
}
\end{center}
\caption{Example prompt.}
\label{fig:prompt}
\end{figure*}

\begin{figure*}[h]
\begin{center}
{\scriptsize
\rule{\textwidth}{1.2pt}
\begin{verbatim}
system:
<PROBLEM>
What is the arithmetic mean of all of the positive two-digit integers with the property that the integer is equal to the sum 
of its first digit plus its second digit plus the product of its two digits?
</PROBLEM>
\end{verbatim}
\rule{\textwidth}{0.4pt}
\begin{verbatim}
assistant:
<THINK>
Let's first try to understand the problem statement. We need to find the arithmetic mean of all two-digit integers that 
satisfy the following condition:
- The integer is equal to the sum of its first digit and its second digit and the product of its two digits.

Let's say the two-digit integer is represented as "ab", where a and b are the first and second digits, respectively. Then, we 
can write the condition as:
10a + b = a + b + ab
Simplifying this equation, we get:
9a = ab
b = 9

So, we see that b must be 9. Therefore, the two-digit integer can be represented as "a9". Substituting this in the original 
equation, we get:
10a + 9 = a + 9 + 9a
Simplifying this equation, we get:
a = 1

So, the two-digit integers that satisfy the given condition are 19, 29, ..., 99. There are 9 such integers. Therefore, the 
arithmetic mean of these integers is:
(19 + 29 + ... + 99) / 9 = 54

Therefore, the answer is 54.
</THINK>

<PYTHON>
# Let's verify the answer using Python code

# Find all two-digit integers that satisfy the given condition
nums = []
for i in range(10, 100):
    a, b = divmod(i, 10)
    if 10*a + b == a + b + a*b:
        nums.append(i)

# Calculate the arithmetic mean
mean = sum(nums) / len(nums)

# Output the answer
print(mean)
</PYTHON>
\end{verbatim}
\rule{\textwidth}{0.4pt}
\begin{verbatim}
system:
<OUTPUT tool="PYTHON">
59.0
</OUTPUT>
\end{verbatim}
\rule{\textwidth}{0.4pt}
\begin{verbatim}
assistant:
<THINK>
It seems like there is a discrepancy between the answer I calculated using the THINK tag and the answer returned by the 
system using the PYTHON tag. Let me check my calculations again.

The two-digit integers that satisfy the given condition are 19, 29, ..., 99. There are 9 such integers. Therefore, the
arithmetic mean of these integers is:
(19 + 29 + ... + 99) / 9 = (9/2)(19 + 99) / 9 = 59

So, my initial answer was incorrect. The correct answer is 59.
</THINK>

<ANSWER>
The arithmetic mean of all of the positive two-digit integers with the property that the integer is equal to the sum of its
first digit plus its second digit plus the product of its two digits is 59.
</ANSWER>
\end{verbatim}
\rule{\textwidth}{1.2pt}
}
\end{center}
\caption{The LLM makes a computational error within the CoT, but corrects the error using the Python results and reaches the correct answer}
\label{fig:chat1}
\end{figure*}

\begin{figure*}[h]
\begin{center}
{\scriptsize
\rule{\textwidth}{1.2pt}
\begin{verbatim}
system:
<PROBLEM>
Compute
\[\sum_{1 \le j < i} \frac{1}{2^{i + j}},\]where the sum is taken over all positive integers $i$ and $j$ such that
$1 \le j < i.$
</PROBLEM>
\end{verbatim}
\rule{\textwidth}{0.4pt}
\begin{verbatim}
assistant:
<THINK>
We can write out the first few terms of the sum to see if we can spot a pattern:
\begin{align*}
\frac{1}{2^2} + \frac{1}{2^3} + \frac{1}{2^4} + \frac{1}{2^5} + \dotsb &= \frac{1}{2^2} \cdot \left(1 + \frac{1}{2} + \frac{1}
{2^2} + \frac{1}{2^3} + \dotsb\right) \\
&= \frac{1}{2^2} \cdot \frac{1}{1 - 1/2} \cdot \frac{1}{2} \\
&= \frac{1}{4}.
\end{align*}
Similarly, we can write out the terms for $i = 3,$ $i = 4,$ and so on, and we get
\[\frac{1}{2^3} + \frac{1}{2^4} + \frac{1}{2^5} + \dotsb = \frac{1}{8},\]
\[\frac{1}{2^4} + \frac{1}{2^5} + \frac{1}{2^6} + \dotsb = \frac{1}{16},\]
and so on. Thus, the sum we want to compute is
\[\frac{1}{4} + \frac{1}{8} + \frac{1}{16} + \dotsb.\]
This is a geometric series with first term $\frac{1}{4}$ and common ratio $\frac{1}{2},$ so its sum is
\[\frac{\frac{1}{4}}{1 - \frac{1}{2}} = \boxed{\frac{1}{2}}.\]
</THINK>

<PYTHON>
sum = 0
for i in range(2, 100):
    for j in range(1, i):
        sum += 1 / 2**(i + j)
print(sum)
</PYTHON>
\end{verbatim}
\rule{\textwidth}{0.4pt}
\begin{verbatim}
system:
<OUTPUT tool="PYTHON">
0.33333333333333326
</OUTPUT>
\end{verbatim}
\rule{\textwidth}{0.4pt}
\begin{verbatim}
assistant:
<THINK>
Our answer from the Python code is not the same as the answer we got from our previous calculations. Let's try to figure out 
which one is correct.

We can see that the sum we want to compute is
\[\sum_{i = 2}^\infty \sum_{j = 1}^{i - 1} \frac{1}{2^{i + j}}.\]
We can switch the order of summation to get
\[\sum_{j = 1}^\infty \sum_{i = j + 1}^\infty \frac{1}{2^{i + j}}.\]
Then, we can evaluate the inner sum:
\begin{align*}
\sum_{i = j + 1}^\infty \frac{1}{2^{i + j}} &= \frac{1}{2^{2j + 1}} + \frac{1}{2^{2j + 2}} + \frac{1}{2^{2j + 3}} + \dotsb \\
&= \frac{1}{2^{2j}} \cdot \left(\frac{1}{2} + \frac{1}{2^2} + \frac{1}{2^3} + \dotsb\right) \\
&= \frac{1}{2^{2j + 1}}.
\end{align*}
Thus, our sum is
\begin{align*}
\sum_{j = 1}^\infty \frac{1}{2^{2j + 1}} &= \frac{1}{2^3} + \frac{1}{2^5} + \frac{1}{2^7} + \dotsb \\
&= \frac{1}{8} \cdot \left(1 + \frac{1}{2} + \frac{1}{2^2} + \dotsb\right) \\
&= \frac{1}{8} \cdot \frac{1}{1 - 1/2} \\
&= \frac{1}{4}.
\end{align*}
Therefore, the answer from our Python code is incorrect, and the answer is $\boxed{\frac{1}{4}}.$
</THINK>

<ANSWER>
The sum is $\boxed{\frac{1}{4}}.$
</ANSWER>
\end{verbatim}
\rule{\textwidth}{1.2pt}
}
\end{center}
\caption{The LLM writes the correct Python code but ignores the results and outputs a different answer.}
\label{fig:chat2}
\end{figure*}

\begin{figure*}[h]
\begin{center}
{\scriptsize
\rule{\textwidth}{1.2pt}
\begin{verbatim}
system:
<PROBLEM>
Evaluate $(-125)^{4/3}$.
</PROBLEM>
\end{verbatim}
\vspace{2mm}
\rule{\textwidth}{0.4pt}
\begin{verbatim}
assistant:
<PYTHON>
import math

print((-125)**(4/3))
</PYTHON>
\end{verbatim}
\rule{\textwidth}{0.4pt}
\begin{verbatim}
system:
<OUTPUT tool="PYTHON">
(-312.50000000000017-541.2658773652738j)
</OUTPUT>
\end{verbatim}
\rule{\textwidth}{0.4pt}
\begin{verbatim}
assistant:
<THINK>
We want to evaluate $(-125)^{4/3}$. We can write this as $(-1)^{4/3} \cdot 125^{4/3}$. 

Now, $(-1)^{4/3} = ((-1)^2)^{2/3} = 1^{2/3} = 1$. 

So, we just need to evaluate $125^{4/3}$. We can write this as $(125^{1/3})^4$. 

Using the fact that $125 = 5^3$, we have $125^{1/3} = 5$. 

Therefore, $(-125)^{4/3} = (-1)^{4/3} \cdot 125^{4/3} = 1 \cdot (5)^4 = \boxed{625}$.
</THINK>

<ANSWER>
$(-125)^{4/3} = \boxed{625}$.
</ANSWER>
\end{verbatim}
\rule{\textwidth}{1.2pt}
}
\end{center}
\caption{There is a problem with the Python code, but the CoT has arrived at the correct answer.}
\label{fig:chat3}
\end{figure*}

\end{document}